\documentclass[lettersize,journal]{IEEEtran}
\usepackage{amsmath,amsfonts}
\usepackage{algorithmic}
\usepackage{algorithm}
\usepackage{array}
\usepackage[caption=false,font=normalsize,labelfont=sf,textfont=sf]{subfig}
\usepackage{textcomp}
\usepackage{stfloats}
\usepackage{url}
\usepackage{verbatim}
\usepackage{graphicx}
\usepackage{cite}
\usepackage{pgfplots}
\usepackage{gensymb}
\usepackage{makecell}
\usepackage{booktabs}

\newcommand{\blue}[1]{#1}

\usepackage{listofitems}
\usepackage{xcolor}
\usepackage{tikz}
\tikzset{>=latex} 
\usetikzlibrary{arrows.meta}
\usetikzlibrary{positioning,calc}
\usetikzlibrary{backgrounds}
\usepgfplotslibrary{fillbetween}
\pgfplotsset{compat=1.12}
\usepackage[outline]{contour}
\contourlength{1.4pt}

\colorlet{myred}{red!80!black}
\colorlet{myblue}{blue!80!black}
\colorlet{mygreen}{green!60!black}
\colorlet{myorange}{orange!70!red!60!black}
\colorlet{mydarkred}{red!30!black}
\colorlet{mydarkblue}{blue!40!black}
\colorlet{mydarkgreen}{green!30!black}
\tikzstyle{node}=[thick,circle,draw=myblue,minimum size=22,inner sep=0.5,outer sep=0.6]
\tikzstyle{node in}=[node,green!20!black,draw=mygreen!30!black,fill=mygreen!25]
\tikzstyle{node hidden}=[node,blue!20!black,draw=myblue!30!black,fill=myblue!20]
\tikzstyle{node convol}=[node,orange!20!black,draw=myorange!30!black,fill=myorange!20]
\tikzstyle{node out}=[node,red!20!black,draw=myred!30!black,fill=myred!20]
\tikzstyle{connect}=[thick,mydarkblue] 
\tikzstyle{connect arrow}=[-{Latex[length=4,width=3.5]},thick,mydarkblue,shorten <=0.5,shorten >=1]
\tikzset{ 
  node 1/.style={node in},
  node 2/.style={node hidden},
  node 3/.style={node out},
}
\def\nstyle{int(\lay<\Nnodlen?min(2,\lay):3)} 

\begin{document}

\title{Non-adversarial Robustness of Deep Learning Methods for Computer Vision}

\author{Gorana Gojić, Vladimir Vincan, Ognjen Kundačina, Dragiša Mišković and Dinu Dragan 
\thanks{Gorana Gojić is with the The Institute of Artificial Intelligence Research and Development of Serbia, 1 Fruškogorska, 21000 Novi Sad, Serbia (e-mail:
gorana.gojic@ivi.ac.rs), ORCID ID (https://orcid.org/0000-0002-6100-5871)
 }
\thanks{Vladimir Vincan is with the The Institute of Artificial Intelligence Research and Development of Serbia, 1 Fruškogorska, 21000 Novi Sad, Serbia (e-mail:
vladimir.vincan@ivi.ac.rs), ORCID ID (https://orcid.org/0000-0001-8356-8133)
 }%
\thanks{Ognjen Kundačina is with the The Institute of Artificial Intelligence Research and Development of Serbia, 1 Fruškogorska, 21000 Novi Sad, Serbia (e-mail:
ognjen.kundacina@ivi.ac.rs), ORCID ID (https://orcid.org/0000-0003-0198-3363)
 }%
 \thanks{Dragiša Mišković is with the The Institute of Artificial Intelligence Research and Development of Serbia, 1 Fruškogorska, 21000 Novi Sad, Serbia (e-mail:
dragisa.miskovic@ivi.ac.rs), ORCID ID (https://orcid.org/0000-0002-0455-9552)
}%
 \thanks{Dinu Dragan is with the  The Faculty of Technical Sciences, University of Novi Sad, 6 Trg Dositeja Obradovića, 21000 Novi Sad, Serbia (e-mail:
dinud@uns.ac.rs)
}}%



\maketitle

\begin{abstract}
Non-adversarial robustness, also known as natural robustness, is a property of deep learning models that enables them to maintain performance even when faced with distribution shifts caused by natural variations in data. However, achieving this property is challenging because it is difficult to predict in advance the types of distribution shifts that may occur. To address this challenge, researchers have proposed various approaches, some of which anticipate potential distribution shifts, while others utilize knowledge about the shifts that have already occurred to enhance model generalizability. In this paper, we present a brief overview of the most recent techniques for improving the robustness of computer vision methods, as well as a summary of commonly used robustness benchmark datasets for evaluating the model's performance under data distribution shifts. Finally, we examine the strengths and limitations of the approaches reviewed and identify general trends in deep learning robustness improvement for computer vision. 
\end{abstract}


\section{Introduction}
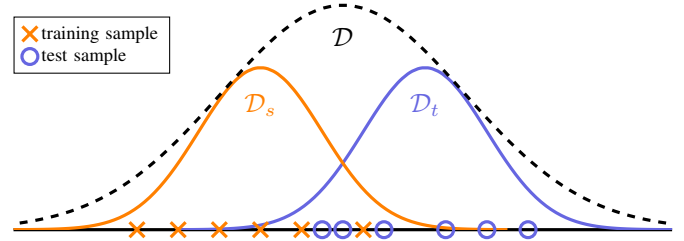
\begin{figure}
    \centering
    \begin{tikzpicture}
        \message{Slika 1^^J}

        \pgfmathdeclarefunction{gauss}{3}{%
            \pgfmathparse{1/(#3*sqrt(2*pi))*exp(-((#1-#2)^2)/(2*#3^2))}%
        }
        \def\N{50}
      
        \def\mean{4};
        \def\dev{3.0};
        \def\xmax{\mean+3.2*\dev};
        \def\ymin{{-0.1*gauss(\mean,\mean,\dev)}};
        \def\h{0.07*gauss(\mean,\mean,\dev)};
        \def\a{\mean-0.8*\dev};
      
        \begin{axis}[
            hide y axis,
            xmin=-16,
            xmax=16,
            ymin=0,
            ymax=0.35,
            axis x line=middle,
            x axis line style={
                very thick, 
                -},
            ticks=none,
            width=0.57*\textwidth, 
            height=0.4*\textwidth,
            legend cell align={left},
            legend style={
                at={(0,0.5)},
                anchor=north west,
                nodes={scale=0.7},},
        ]
        
            \addplot+[mark=none,samples=\N,myblue!60,very thick,name path=training,smooth,forget plot][domain=\mean-4*\dev:\mean+4*\dev] {gauss(x,\mean,\dev)};
            \addplot+[mark=none,samples=\N,orange,very thick,name path=test,smooth,forget plot][domain=-\mean-4*\dev:-\mean+4*\dev] {gauss(x,-\mean,\dev)};
            \addplot+[mark=none,samples=\N,very thick,black,dashed,name path=test,smooth,forget plot][domain=-6*\dev:+6*\dev] {250*gauss(x,0,2*\dev)*gauss(x,0,12*\dev)};
    
            \addplot+[only marks, very thick, orange, mark=x, mark size=4pt] table {
                -10 0 
                -8 0
                -6 0
                -4 0
                -2 0 
                1 0
            };
            \addplot+[only marks, very thick, myblue!60, mark=o, mark size=3pt] table {
                -1 0
                0 0 
                2 0
                5 0
                7 0
                9 0
            };
    
            \node[myblue!60,below,very thick] at ({\mean},{0.9*gauss(\mean,\mean,\dev)}) {$\mathcal{D}_t$};
            \node[orange,below,very thick] at ({-\mean},{0.9*gauss(\mean,\mean,\dev)}) {$\mathcal{D}_s$};
            \node[below] at (0,{1.3*gauss(0,0,\dev)}) {$\mathcal{D}$};
    
            \addlegendentry{training sample}
            \addlegendentry{test sample}
        
        \end{axis}
    \end{tikzpicture}
    \caption{Distribution shift illustration. A priori unknown real-world data distribution $\mathcal{D}$ (denoted by dashed line) with source $\mathcal{D}_{s}$ and target distribution $\mathcal{D}_{t}$ (denoted with orange and blue solid lines) being shifted relative to $\mathcal{D}_{s}$.}
    \label{fig:distribution shift}
\end{figure}

Robustness is an important property of deep learning models. It refers to the model's ability to produce expected outputs in cases when input data differs from the data the model has been trained on \cite{drenkow_2021_robustness}. Literature distinguishes between adversarial and non-adversarial robustness also known as natural robustness. Adversarial robustness refers to the model's ability to accurately classify input data that has been intentionally corrupted in a way that aims to fool the model into making an incorrect prediction with high confidence. This type of robustness relies on exploiting model and data properties \cite{goodfellow_2014_adversarial}. While adversarial robustness prevents the model malfunction under intentional attacks \cite{goodfellow_2014_adversarial}, non-adversarial robustness preserves the model performance under naturally-induced data transformations such as environmental and sensory transformations \cite{drenkow_2021_robustness}. Natural robustness is a desirable model property, as deployed models often encounter test data that differs from training data. For example, a model trained to recognize traffic signs on images collected in an area where rarely snows might generate an unpredictable output for a test image displaying the same sign under severe snowing conditions \cite{goodfellow_2014_adversarial}. Differences in training and test datasets stem from distribution shifts that are common in practice due to the changing nature of data distributions, which can be both temporal or non-temporal (i.e., changes across locations, camera choice, image artifacts) \cite{sugiyama_2012_machine}. In a standardized machine learning pipeline, a training dataset is collected by sampling from theoretically infinite, high-dimensional data space with a priori unknown distribution $\mathcal{D}$. A set of drawn samples dictates a source distribution of a training dataset, denoted with an orange curve in Fig. \ref{fig:distribution shift}, that the model learns through training. Trained model inferences on test data having a target distribution, denoted by a blue curve in Fig. \ref{fig:distribution shift},  shifted relative to the source distribution. 
Identifying the causes of distribution shift can be straightforward in some cases, such as in \cite{hendrycks_2018_benchmarking}, where a model trained on sharp images was tested on blurry images that were not present in the training set. However, in many cases, the causes for distribution shift are not obvious and cannot be easily identified \cite{recht_2019_imagenet}. In \cite{recht_2019_imagenet} the authors acquire new test datasets for CIFAR-10 \cite{cifar10-dataset} and ImageNet \cite{imagenet-dataset} following dataset acquisition guidelines and observe significantly decreased accuracy on a broad range of models when compared to original datasets. Assessing the causes of distribution shifts accurately, especially in high-dimensional data, poses a challenge to training models that can generalize well to an arbitrary sample from distribution $\mathcal{D}$ \cite{goodfellow_2016_dl}.

When a model performs well on instances from a different distribution compared the the one it was trained on, it demonstrates out-of-distribution (OOD) robustness. If the model performs consistently well on instances from the same distribution, it demonstrates in-distribution (ID) robustness, ensuring stable performance on unseen data from the same distribution. While ID robustness does not explore the model's generalization under distribution shifts, it is used as a standard method to detect overfitting and underfitting during the training. It is also used as a control metric for methods boosting OOD generalization, since it has been shown that some of the methods can increase OOD while simultaneously hurt model performance ID \cite{kumar_2022_finetuning}.

In this paper, we overview recent advancements in non-adversarial robustness, or shorter robustness, as we will refer to in this paper. While there are many papers addressing well-explored adversarial robustness \cite{croce_2020_advers1, cohen_2019_adver2, madry_2018_adver3}, non-adversarial robustness has received significantly less attention. Robustness improvement techniques have been proposed and discussed in multiple papers in the context of proposing a novel robustness benchmark dataset \cite{hendrycks_2018_benchmarking}, robustness method \cite{hendrycks_2019_dg_augmix, yun_2019_dg_aug_cutmix, devries_2017_dg_aug_cutout}, or both \cite{hendrycks_2021_deepaugment}. However, we were unable to find a systematic overview of these methods that could serve as an entry point to the non-adversarial robustness research field. Thus, this paper gives a broad overview of robustness improvement methods based on domain adaptation and domain generalization approaches. We discuss the advantages and disadvantages of the approaches and complement the study with an overview of publicly available robustness benchmark datasets in computer vision.

The paper is organized as follows. Section \ref{sec:deep-learning-fundamentals} gives a brief overview of the main deep learning principles that might help in following the content of Sections \ref{sec:domain-generalization} and \ref{sec:domain-adaptation} where we discuss major robustness improvement methodologies in domain generalization and domain adaptation fields. In Section \ref{sec:benchmarking-datasets} we list major publicly available robustness benchmarking datasets, discuss the findings in Section \ref{sec:discussion}, and conclude in Section \ref{sec:conclusions}.

\section{Fundamentals of Deep Learning}
\label{sec:deep-learning-fundamentals}

The development of deep learning was motivated by the failure of traditional machine algorithms to adequately solve complex tasks such as speech or object recognition. These tasks rely on processing high-dimensional data with an immense input data space. With just a subset of the data points from the entire space, traditional machine learning algorithms could not provide good generalization on unseen data points, since they rely on the similarity between the new data points and those they have been previously learned on \cite{goodfellow_2016_dl}. To address this issue, deep learning algorithms are designed as complex functions that use sparsely filled input data space to learn how to generalize by incorporating knowledge about the underlying data distribution. Here we introduce the basic concepts of feed-forward deep neural networks (DNNs) since the majority of literature studying robustness relies on DNNs. We explain the fully-connected neural network (FCNN) that serves as a base for other DNN specializations, such as convolutional neural networks (CNNs). Since CNNs are widely recognized as the primary architecture for computer vision tasks, we also provide a brief explanation of their structure.

\subsection{Fully-connected neural networks}
\begin{figure}
    \centering
    \begin{tikzpicture}[
        x=2.7cm,y=1.6cm,
        node in/.style={node,black,draw=white,fill=white}, 
        node out/.style={node,white,draw=white,fill=white},
        scale=0.88, transform shape
    ]
        \message{^^JSlika 2}
        \def\NI{3} 
        \def\NO{1} 
        \def\yshift{0.4} 
      
        \foreach \i [evaluate={
            \c=int(\i==\NI); 
            \y=\NI/2-\i-\c*\yshift; 
            \index=(\i<\NI?int(\i):"n");}]
        in {1,...,\NI}{ 
            \node[node in,outer sep=0.6] (NI-\i) at (0,\y) {$x_{\index}$};
            \message{in: \y, }
        }
      
        \foreach \i [evaluate={
            \c=int(\i==\NO); 
            \y=\NO/2-\i-\c*\yshift; 
            \index=(\i<\NO?int(\i):"m");}]
        in {\NO,...,1}{ 
            \message{out: \y }
            \node[node hidden] (NO-\i) at (1,\y) {$\sum$};
            \foreach \j [evaluate={\index=(\j<\NI?int(\j):"n");}] in {1,...,\NI}{ 
                \draw[connect arrow,white,line width=1.2] (NI-\j) -- (NO-\i);
                \draw[connect arrow] (NI-\j) -- (NO-\i) node[pos=0.50] {\contour{white}{$w_\index$}};
            }
        }
        
        \node[node hidden] (N+) at (1.7,-0.9) {+};
        \node[node in] (Nb) at (1.7,0) {b};
        \node[node out] (Nout) at (2.4,-0.9) {out};
        \draw[connect arrow] (NO-1) -- (N+);
        \draw[connect arrow] (Nb) -- (N+);
        \draw[connect arrow] (N+) -- (Nout);
      
        \path (NI-\NI) --++ (0,1+\yshift) node[midway,scale=1.2] {$\vdots$};
        
        \node[right=-0.15,scale=0.95] at (Nout)
            {$\begin{aligned}
                &activation(\sum_{i=1}^nx_iw_i+b)
            \end{aligned}$};

    \end{tikzpicture}
    \caption{The structure of a neuron located in the hidden layer of a FCNN.}
    \label{fig:neuron}
\end{figure}
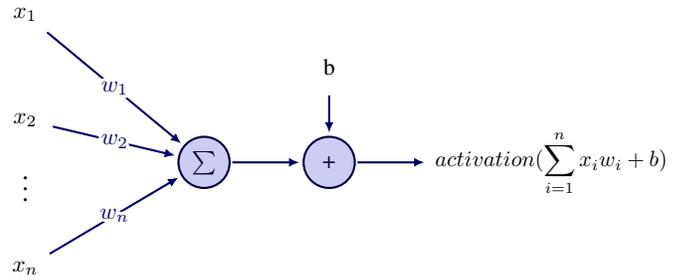

The goal of DNN is to learn a function approximation $y'=f(x; \theta)$ that maps the input $x$ to the prediction $y'$ ideally equal to expected output $y$ by fitting the function parameters $\theta$ on some input dataset $D=\{(x_{1}, y_{1}), (x_{2}, y_{2}), \ldots, (x_{k}, y_{k})\}$ with $k$ data points. The learning is performed by computational units called neurons with the general architecture shown in Fig. \ref{fig:neuron}. Each neuron can receive multiple inputs $x_{i}$ weighted by a set of learnable parameters $w_{i}$ and learnable bias $b$ to produce a single scalar value representing the neuron's activation. Since neurons are inherently linear, a nonlinear activation function is applied to the scalar to support DNNs in learning nonlinear transformations. The activation of one neuron is propagated to succeeding neurons to facilitate the learning of complex feature representations from the data \cite{goodfellow_2016_dl}.

\begin{figure}
    \centering
    \begin{tikzpicture}[
        scale=0.8,
        x=2.2cm,y=1.4cm,
        node in/.style={node,white,draw=white,fill=white},
        node out/.style={node,white,draw=white,fill=white},
    ]
      \message{^^J FCNN}
      \readlist\Nnod{4,4,4,2,2}
    
      \message{^^J  Layer}
    
      \draw[myblue!40,fill=myblue,fill opacity=0.02,rounded corners=2]
        (1.7, 1.5) rectangle++ (0.6, -4);
      \node[below=0.1,align=center,black] at (2,-2.5) {$\mathcal{L}_{in}$};
      \draw[myblue!40,fill=myblue,fill opacity=0.02,rounded corners=2]
        (2.7, 1.5) rectangle++ (0.6, -4);
      \node[below=0.1,align=center,black] at (3,-2.5) {$\mathcal{L}_{hidden}$};
      \draw[myblue!40,fill=myblue,fill opacity=0.02,rounded corners=2]
        (3.7, 0.5) rectangle++ (0.6, -2);
      \node[below=0.1,align=center,black] at (4,-2.5) {$\mathcal{L}_{out}$};
      
      \foreachitem \N \in \Nnod{ 
        \edef\lay{\Ncnt} 
        \message{^^J \lay,}
    
        \pgfmathsetmacro\prev{int(\Ncnt-1)} 
        \foreach \i [evaluate={\y=\N/2-\i; \x=\lay; \n=\nstyle;}] in {1,...,\N}{ 
          
          \node[node \n] (N\lay-\i) at (\x,\y) {$n_\i$};
          
            \ifnum\lay>1 
                \ifnum\lay=2
                    \draw[connect arrow] (N\prev-\i) -- (N\lay-\i);
                \else
                    \ifnum\lay=5
                        \draw[connect arrow] (N\prev-\i) -- (N\lay-\i);
                    \else
                        \foreach \j in {1,...,\Nnod[\prev]}{ 
                            \draw[connect arrow] (N\prev-\j) -- (N\lay-\i); 
                        }
                    \fi
                \fi
            \fi 
        }
      }
    \end{tikzpicture}
    \caption{A FCNN consisting of the input layer $\mathcal{L}_{in}$,  the hidden layer $\mathcal{L}_{hidden}$, and the output layer $\mathcal{L}_{out}$.}
    \label{fig:deep-neural-network}
\end{figure}
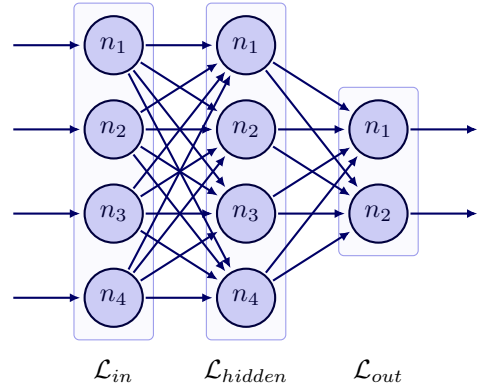

Neurons are organized into layers as shown in Fig. \ref{fig:deep-neural-network}. The number of layers determines the depth of the network, and it is characteristic of DNNs to have many layers \cite{he_2016_deep_resnets, huang_2017_densenet, tan_2019_efficientnet, dosovitskiy_2020_vit}. The first layer in the network is the input layer, which serves as data entry point. It is followed by hidden layers that usually contain many neurons simultaneously calculating their activations based on the output of the preceding layer. The final, output layer is responsible for generating DNN prediction for a given task. 

The process of fitting DNN parameters on the training dataset $D_{train}$ is called training. The scope of the training pertains to the adjustable weights $\mathbf{W}$ and biases $\mathbf{b}$ attributed to all trainable neurons in the network, where the weight vector $w_{i}^{j}$ and bias $b_{i}^{j}$ correspond to a neuron $i \in [1,n]$ located in layer $j \in [1,m]$. The training is formulated as an optimization problem over a parameter set $\theta$ consisting of $\mathbf{W}$ and $\mathbf{b}$ that minimizes a loss function $L$ serving as an optimization objective function. It is typically performed following a two-pass back-propagation algorithm \cite{rumelhart_1986_backprop}. In the first, feed-forward pass, the loss value is calculated to estimate the error between DNN predictions and desired outputs for training samples from $D_{train}$. The gradients over $\theta$ are then calculated to minimize the loss using some variation of gradient-based algorithms, i.e., standard stochastic gradient descent (SGD), SGD with momentum, Adagrad \cite{duchi_2011_adagrad}, RMSProp \cite{tijmen_2012_rmsprop}, or Adam \cite{kingma_2015_adam}. Next, gradients are used to update $\theta$. The process is iteratively repeated until the stop condition is met, i.e., exceeded number of iterations or gradient descent learning rate being lower than the predefined threshold value. Hyperparameters are the values used to control the training process and define DNN architecture. They can be adjusted manually or automatically \cite{bergstra_2013_hyperparam, hospedales_2022_metalearning} to determine the optimal configuration for a certain task. In traditional DNN they can be immutable (e.g., a choice of a loss function, number of DNN layers) or mutable (e.g., learning rate in gradient descent). In meta-learning paradigm, some immutable hyperparameters can be treated as mutable and learned by another algorithm \cite{hospedales_2022_metalearning}.

In cases when DNN is not trained on extremely large datasets, it is advisable to introduce regularization to prevent the network from overfitting \cite{goodfellow_2016_dl}. The overfitted model has a poor ID and OOD robustness since it has not learned patterns in data, but has memorized training data instead. Regularization can be imposed on the training data, neurons, and training algorithm. Data augmentation regularizes the network by enlarging the training dataset and diversifying source distribution (see Section \ref{sec:domain-adaptation} for details). Dropout \cite{srivastava_2014_dropout} prevents the network from favoring certain neurons in decision-making by randomly eliminating their activations during the training. Regularization strategies, such as L1 and L2 regularization \cite{goodfellow_2016_dl}, constrain loss function by imposing restrictions on weight values. Another standard regularization strategy is to stop training when a training error is consistently decreasing while the validation error calculated on a control group of unseen data samples is consistently increasing. The output of a training stage is a DNN model evaluating function $y'$ with fixed $\theta$. The model is used to generate predictions for unseen data samples that come from the test dataset $D_{test}$ in a process called inference. 

\subsection{Convolutional neural networks}
\blue{
Convolutional neural networks (CNNs) \cite{lecun_1998_cnn} are a type of DNNs designed to handle data with a topological grid structure, such as images. While the input of the FCNN neuron is densely connected to the outputs of all neurons from the previous layer, the CNN neuron is sparsely connected to activations in the previous layer. Thus CNNs exhibit locality inductive bias, meaning that by design CNN neurons extract features from spatially close grid elements in input data. The bias is motivated by the assumption that meaningful features in the grid are assembled of spatially close pixels \cite{goodfellow_2016_dl}. Each CNN layer learns sets of trainable kernels which are weights shared among the neurons in the layers. Weight sharing is another inductive bias in CNNs that enables them to detect similar features in a translation-invariant manner. Learning features in subsequent layers enables CNNs to learn low-level features from input data, which are then used to gradually assemble more complex features. 

CNNs have been successfully utilized in many tasks such as object detection} \cite{he_2016_deep_resnets, girshick_2014_rcnn}, classification \cite{he_2016_deep_resnets, huang_2017_densenet}, \cite{he_2016_deep_resnets}, semantic segmentation \cite{ronneberger_2015_unet, girshick_2014_rcnn}, instance segmentation \cite{cheng_2021_instancesegm}, and representation learning \cite{radford_2015_dcgan}. Nowadays modern CNN architectures have many stacked convolutional layers that may be followed by fully connected layers \cite{he_2016_deep_resnets, huang_2017_densenet, girshick_2014_rcnn} or can act as feature extractors \cite{radford_2015_dcgan}. It is important to note that vision transformers \cite{dosovitskiy_2021_vit} have emerged recently as a successful alternative to CNNs in image processing. However, due to review limitations, we omit them from the study.

\section{Domain Generalization Techniques}
\label{sec:domain-generalization}
Domain generalization is a machine learning field that aims to train the model on source distribution data so that it can generalize well to arbitrary target distribution. The unexpected nature of target distribution makes data generalization a challenging task. One approach to address the task is to learn feature representations that are distribution shift invariant. Further in this section, we discuss some implementations of data generalization approach.

\subsection{Data augmentation}
One approach to anticipate distribution shifts is to increase source distribution diversity and simultaneously enlarge the dataset by applying transformations on data sample copies \cite{goodfellow_2016_dl}. The idea behind data augmentation is to support the model generalization by learning feature representations invariant to data augmentations. Simple geometric image augmentations, such as random rotation, translation, and mirroring, do not change data distribution significantly, but can still have a regularization effect. Although simple to implement, some of these augmentations can lead to information loss (i.e., rotating the image for 45$\degree$) \cite{yang_2022_augmentations}. In \cite{devries_2017_dg_aug_cutout, zhong_2017_dg_aug_erase} authors propose augmentations based on erasing image patches. The underlying idea is to increase robustness against occlusions by erasing image patches. In this way, the network learns not to heavily rely on some specific data feature when generating predictions. A cutout augmentation \cite{devries_2017_dg_aug_cutout} implements this conceptually similar to dropout, by setting to zero activations of input neurons associated with randomly selected input image patch. While dropout targets individual neurons in hidden layers, cutout zeroes out weights of spatially connected groups of input layer neurons.
The authors of \cite{zhong_2017_dg_aug_erase} propose conceptually similar approach to that in \cite{devries_2017_dg_aug_cutout}, but with extensive experiments on patch size and erased pixel color. Data augmentation by erasing is intended to improve generalization under occlusions. Although this technique may have limited benefits when occlusions are not present in the target dataset, it can still provide additional advantages when combined with other data augmentation methods \cite{zhong_2017_dg_aug_erase}. An approach in \cite{yun_2019_cutmix} argues that replacing image patches with semantically unmeaningful values introduces information loss and proposes an approach where the patch that has been cut out is replaced with a patch from another image in the training dataset. The augmentation demonstrates OOD robustness by extending source distribution with new training samples generated as a combination of two existing ones, combining both raw images and corresponding labels as initially proposed in \cite{zhang_2018_mixup}. The basic concern in synthetic augmentation techniques is that creating semantically non-meaningful combinations will map poorly to real-world data \cite{taori_2020_measuring}, e.g., it is improbable that a real-world image will display a horse with a frog's head patch. Approaches in  \cite{hendrycks_2019_dg_augmix} and \cite{hendrycks_2021_deepaugment} address this issue by proposing complex, yet OOD robust augmentations that do not significantly change augmented image statistics. In \cite{hendrycks_2019_dg_augmix} authors propose chaining many simple augmentations to create a complex augmentation. The augmentation pipeline starts with randomly sampling augmentations from a predefined set into $n$ disjunct subsets and sequentially applying each augmentation subset in parallel starting from the raw image. The pipeline outputs multiple augmented versions of the raw image linearly combined with the raw image to obtain the final augmentation. In \cite{hendrycks_2021_deepaugment}, authors propose an image-to-image network that applies random augmentations on network weights, producing uniquely augmented images. A style transfer variant of generative adversarial networks (GAN) \cite{goodfellow_2014_gans} proposed in \cite{taori_2020_measuring} has been shown to facilitate OOD robustness when used as an augmentation technique \cite{hendrycks_2021_deepaugment}. 

While multiple studies emphasize the necessity of combining multiple augmentations to achieve good generalization \cite{devries_2017_dg_aug_cutout, yang_2022_augmentations, hendrycks_2021_deepaugment}, it is a non-trivial task to choose the optimal combination. Theoretical foundations of data augmentation are yet to be explored, and the majority of discoveries in the field are experimental \cite{yang_2022_augmentations}. At the moment, studies suggest that optimal data augmentation choice is specific for a training dataset and task combination \cite{yang_2022_augmentations}, with inappropriate choice leading to reduced model generalization such as in \cite{hendrycks_2021_deepaugment} for DeepFashion Remixed benchmark dataset and training instability \cite{hendrycks_2018_benchmarking} even when applied augmentations seem to be visually almost unnoticeable. 

One proposed solution to this issue is to use algorithms for automated augmentation selection. These algorithms search the augmentation space to find the optimal augmentation policy \cite{cubuk_2019_autoaugment, lim_2019_fastautoaugment, ho_2019_population, cubuk_2020_randaugment}. The algorithms are built on the assumption that different data features require different augmentations to properly regularize the trained model. However, automatic augmentation may not be fully utilized in domain generalization since target distribution is not known in training time. Still, the approach might have benefits in supervised domain adaptation (refer to Section \ref{sec:domain-adaptation}) where model is adapted to target distribution.

\subsection{DNN architecture-based techniques} 
Although data augmentation tends to be the major robustness improvement technique, changes in DNN architecture can also be used to increase OOD natural robustness. 

\noindent\textbf{Network depth.} In \cite{hendrycks_2018_benchmarking} it is demonstrated that deeper versions of DenseNet \cite{huang_2018_densenets} and ResNeXt \cite{xie_2017_resnexe} achieve higher robustness on ImageNet-C and ImageNet-P synthetic benchmarks when comparing to their shallower counterparts. Similar conclusions are reached in \cite{xie_2019_intriguing} but for adversarial robustness. Although we do not intend to review adversarial robustness techniques, we include the study to denote that OOD robustness increases with network depth in both adversarial and non-adversarial scenarios to support findings in \cite{hendrycks_2018_benchmarking} in absence of other supporting non-adversarial OOD studies. The results from \cite{taori_2020_measuring} suggest that natural robustness to synthetic distribution shifts does not apply to real-world distribution shifts, unless the model is trained on large and varied datasets. As a result, the conclusions drawn in \cite{hendrycks_2018_benchmarking} only apply to synthetic distribution shifts. However, succeeding work in \cite{hendrycks_2021_deepaugment} contradicts \cite{taori_2020_measuring} by achieving comparable robustness both on synthetic and real-world benchmarks.

\noindent\textbf{Network architecture.} It has been demonstrated that some architectures exhibit increased natural robustness \cite{huang_2018_densenets, xie_2017_resnexe, ke_2017_multigrid, huang_2018_multiscale} when compared to base ResNet architecture. DenseNet and ResNeXt are shown to be more robust both on perturbations and corruptions \cite{hendrycks_2018_benchmarking}. The increased robustness results from improved data aggregation capabilities introduced through enriched connections between layers. For example, in a DenseNet every convolutional layer inside a dense block uses feature maps from all preceding convolution layers alongside the standard feature map produced by sequential processing. According to \cite{hendrycks_2018_benchmarking}, multi-scale networks \cite{ke_2017_multigrid, huang_2018_multiscale} demonstrate corruption robustness, with no clear benefits on perturbation robustness. Allegedly, this is due to network design where data in different scales is affected differently by corruptions to produce the final prediction. 

Although increasing DNN depth and adapting architecture toward feature aggregation positively influence the model's overall robustness, using these two techniques jointly has demonstrated the best performance.

\section{Domain Adaptation Techniques}
\label{sec:domain-adaptation}
When deploying a DL model in real-world scenarios, data distribution shifts are expected to occur at some point in time, causing the model's performance to deteriorate. Domain adaptation is a field of machine learning aiming to modify a model trained on a source distribution to adapt to data from a target distribution \cite{csurka_2017_domain_adaptation}. In this section we will discuss domain adaptation implementations using transfer \cite{pan_2010_transferlearning}, meta  \cite{hospedales_2022_metalearning}, and few-shot learning \cite{wang_2020_fewshotlearning}.

\subsection{Transfer learning}
\label{sec:da-transfer-learning}

Transfer learning is a machine learning field researching how to utilize knowledge learned in one task to improve another, related task. The basic idea of transfer learning is that DNN learns feature representations gradually from simple, task-agnostic features (e.g., lines) to complex, task-specific features (e.g., nose or ears for face recognition task) that can be fully or partially transferred to another problem. A common way to implement transfer learning is to pretrain a DNN on a large and diverse dataset and then use the first $n$ pretrained layers as an initialization for a new DNN that is then trained on a new dataset. There are two primary transfer learning approaches: fine-tuning and linear probing. With linear probing, the convolutional layer weights of a DNN that act as feature extractors remain fixed, while only the downstream task prediction layers are trained. Fine-tuning, on the other hand, involves updating both the feature extractor and prediction layers.



Using transfer learning in domain adaptation settings is based on the assumption that distribution invariant feature representations can be learned from source and target distribution data \cite{goodfellow_2016_dl}. 
Although pretraining might not necessarily improve model performance \cite{he_2019_rethinking} as it was previously believed \cite{yosinski_2014_transferable}, it has been demonstrated that in certain settings it can increase the models OOD robustness, both in adversarial and non-adversarial settings  \cite{hendrycks_2019_robustness_pretraining}. However, it is shown that the transfer learning efficiency on domain adaptation is dependent on a choice of transfer learning method \cite{kumar_2022_finetuning} and similarities in source and target datasets \cite{yosinski_2014_transferable}. In \cite{kumar_2022_finetuning}, authors use fine-tuning and linear probing approaches to transfer learning, demonstrating that fine-tuning leads to better ID and worse OOD performance compared to linear probing, resulting in higher OOD and lower ID performance. Observing that fine-tuning can distort pretrained features, the authors in \cite{kumar_2022_finetuning} propose a novel transfer learning approach that combines fine-tuning and linear probing.

The recent breakthrough in self-supervised learning (SSL) for representation learning has inspired a community to combine transfer learning with SSL \cite{sun_2019_unsupervised_da, yu_2020_test-time}. Since SSL does not require explicit supervision to learn representations, this approach can be used for both labeled and unlabeled target data. In SSL a model learns to solve an auxiliary, pretext task to learn rich feature representations, e.g., learning to predict image rotation \cite{feng_2019_sslrotation}, or distinguish between images \cite{chen_2020_simclr, he_2020_moco, caron_2021_dino}. SSL approach in \cite{sun_2019_unsupervised_da} uses unlabeled data to learn shared representations on multiple tasks and multiple datasets. The authors train a feature extractor using multitask learning \cite{caruana_1998_multitask} with one task being a downstream task and other tasks being auxiliary SSL tasks (e.g., predicting rotation and localization). All tasks are trained simultaneously using both source and target distribution data. The approach utilizes learning on multiple tasks and datasets to increase knowledge transfer by reducing the distance between learned tasks \cite{yosinski_2014_transferable} and training simultaneously on source and target distribution datasets. Work in \cite{yu_2020_test-time} extends on \cite{sun_2019_unsupervised_da} to support test-time training in online settings. While \cite{sun_2019_unsupervised_da} adapts the model on target data and uses adapted model weights further in inference, \cite{yu_2020_test-time} starts from the model trained solely on source distribution and adapts its weights online as target distribution data arrives.

\subsection{Other techniques}
\label{sec:da-others}
Here we mention other approaches to domain adaptation that have not yet received as much attention as the simplest, transfer learning approach. Our motivation here is not to present a comprehensive overview of existing methods, but to provide the reader with a starting point for further research.

\noindent\textbf{Meta-learning} \cite{hospedales_2022_metalearning}, or learning to learn, is an approach where the meta-learner algorithm learns parts of the base-learner algorithm that performs a downstream task. Some examples include learning an optimization algorithm \cite{ravi_2017_optimization} and base-learner parameter initialization \cite{finn_2017_metalearning_da}. A broad idea behind meta-learning is to support quick and efficient adaptation of the base-learner through meta-training across different tasks and domains. In \cite{li_2020_metalearning_da} the authors propose a meta-learning framework to learn the initial conditions of the existing domain adaptation algorithms relying on the gradient descent method. The suggested framework improves the performance of various domain adaptation techniques in multi-source and semi-supervised scenarios for the target distribution.

\noindent\textbf{Few-Shot Learning.} In few-shot learning (FSL)\cite{wang_2020_fewshotlearning} the task is to train a network to recognize new classes having just a few labeled instances per class. Having a dataset with  few instances per class is insufficient to train DNN from scratch or adapt to target distribution using transfer learning. Therefore, domain adaptation in FSL is often formulated as meta-learning problem, where instances from target distribution are used to speed up convergence of the base-learner. The challenge in few-shot learning domain adaptation is to adapt to new classes that can partially overlap with base classes \cite{saito_2020_universal} or can be completely disjoint \cite{tseng_2020_crossdomain_fsl_da, guo_2020_broader}.

\section{Benchmarking Datasets}
\label{sec:benchmarking-datasets}
In this section, we cover benchmarking datasets used to assess model performance under domain generalization and adaptation approaches. 

\noindent\textbf{Domain generalization benchmarks.}
There are multiple publicly available robustness benchmarks intended to test for the natural robustness of computer vision DNNs. Both synthetic and real-world benchmarks exist. While synthetic benchmarks provide testing in controlled conditions under known distribution shifts, recent work shows that demonstrating robustness on synthetic benchmarks does not necessarily guarantee robustness on real-world distribution shifts \cite{taori_2020_measuring}. To alleviate the issue, real-world benchmarks have been introduces.



\textit{ImageNet-C}\cite{hendrycks_2018_benchmarking} and \textit{ImageNet-P} \cite{hendrycks_2018_benchmarking} are synthetic benchmarks  that decouple robustness benchmarking to corruption and perturbation robustness respectively. To produce both benchmarks, image transformations are applied to original images from the ImageNet dataset. Corruptions are designed to significantly change image statistics and enforce testing in OOD settings. There are 15 types of corruption transformations selected from noise, blur, weather, and digital categories, that are applied to original images with five severity levels to control the degree of image distribution shift. Perturbations are subtle transformations of original images sampled from the same categories as corruptions, but harder to perceive visually when compared to corruptions. The motivation behind the perturbation benchmark is to test subtle data distribution shifts.

Collecting a comprehensive real-world robustness benchmark with systematic distribution shifts can be challenging compared to synthetic benchmarks. The nature of the distribution shifts is difficult to determine due to multiple dimensions of variation that can occur in real-world data simultaneously. For example, an image displaying the same object can be captured with different cameras, from different viewpoints, in different locations, showing the object in various weather conditions. However, the issue raised in \cite{taori_2020_measuring} emphasizes the importance of real-world benchmarking datasets. Here we mention ImageNetV2 \cite{recht_2019_imagenet}, ImageNet-Renditions \cite{hendrycks_2021_deepaugment}, StreetView StoreFronts \cite{hendrycks_2021_deepaugment}, DeepFashionRemixed \cite{hendrycks_2021_deepaugment}, and Real Blurry Images \cite{hendrycks_2021_deepaugment}.

\textit{ImageNetV2} \cite{recht_2019_imagenet} is created by replicating the original ImageNet data collection procces. For this dataset, distribution shifts cannot be clearly identified, but study \cite{recht_2019_imagenet} has shown degraded performance for a range of classifiers trained on ImageNet and tested on ImageNetV2, indicating clear existence of natural distribution shifts in the ImageNetV2 dataset. Similar is true for \textit{ImageNet-Renditions} (ImageNet-R) \cite{hendrycks_2021_deepaugment} which contains image renditions such as paintings, sculptures, and embroidery for the ImageNet classes. The dataset contains images of significantly different textures and local image statistics, shifting ImageNet-R distribution relative to ImageNet. \textit{StreetView StoreFronts} (SVSF) \cite{hendrycks_2021_deepaugment} dataset contains business storefront images. The dataset varies in location, the year the image is captured, and camera properties introducing sensor-induced distribution shifts. Close to the idea of quantifying distribution shifts as introduced in synthetic benchmarks, real-world benchmark \textit{DeepFashion Remixed (DFR)} \cite{hendrycks_2021_deepaugment} uses metadata from DeepFashion2 \cite{ge_2019_deepfashion2} to make systematic changes in object occlusion, zoom, orientation, and scale. Both SVSF and DFR benchmarks provide a way to control distribution shifts by altering dataset parameters (e.g., fixing location for SVSF or varying zoom for DFR and fixing other parameters) and testing DNN OOD generalization with more granularity. \textit{Real Blurry Images} \cite{hendrycks_2021_deepaugment} benchmark contains naturally blurry images. The dataset is proposed as a real-world counterpart of ImageNet-C and ImageNet-P datasets that test model robustness on synthetically generated blurry images. 

\noindent\textbf{Domain adaptation benchmarks.}
\blue{These benchmarks are intended to test how well a given model adapts to changes in data distribution. Thus, domain adaptation benchmarks can be assembled from any two datasets suitable for a particular downstream task. The first in a pair, or a training dataset is of the source distribution, and the second dataset is of the target distribution. While in this setup source and target distributions must be different, literature shows they also should retain some similarities for data distribution shift not to be too extreme. Often, this means having a shared label space \cite{kumar_2022_finetuning, sun_2019_unsupervised_da, shu_2018_dirt}. While domain adaptation benchmarks can be assembled in this manner for any downstream task, in this paper we further discuss domain adaptation benchmarks for object recognition task, as it is a well-established field in deep learning for computer vision.} 

Some of the datasets used to train object recognition DNNs include MNIST \cite{lecun_1998_cnn}, MNIST-M \cite{ganin_2016_da_dataset2}, ImageNet-1K \cite{russakovsky_2015_imagenet}, SVHN \cite{netzer_2011_da_dataset3}, USPS \cite{hull_1994_usps_dataset}, CIFAR-10 \cite{krizhevsky_2009_da_dataset4}, and STL-10 \cite{coates_2011_da_dataset5}. \blue{Some properties of these datasets depicting possible distribution shifts are given in Table \ref{tab:dataset-properties}.} These datasets have been used as domain adaptation benchmarks in \cite{sun_2019_unsupervised_da} by assembling pairs MNIST+MNIST-M, MNIST+SVHN, SVHN+MNIST, USPS+MNIST, CIFAR-10+STL-10, STL-10+CIFAR-10, while in \cite{kumar_2022_finetuning} ImageNet-1K is paired with ImageNet-R \cite{hendrycks_2021_deepaugment}, ImageNet-A \cite{hendrycks_2021_imageneta}, and ImageNetV2 \cite{recht_2019_imagenet} for OOD evaluation. In cases where datasets do not share all labels, differing labels are usually discarded from both datasets \cite{shu_2018_dirt}. There are also specialized benchmarks containing images from multiple domains for all classes in a dataset, with DomainNet \cite{peng_2019_domainnet} being a standard benchmark in the field representing each class in six different domains (e.g., clipart, sketch). Another approach to benchmark domain adaptation is to use domain generalization datasets \cite{yu_2020_test-time}. In few-shot learning (FSL) setup, where label spaces of source and target domains differ, FSL leverages knowledge from the source domain to learn a downstream task on the target domain containing new labels. In \cite{tseng_2020_crossdomain_fsl_da} authors propose a domain adaptation benchmark for non-overlapping label spaces in FSL.

\begin{table*}
\begin{center}
\caption{\blue{Properties of object recognition datasets}}
\label{tab:dataset-properties}
\begin{tabular}{| {c} | {c} | {c} |}
\hline
\textbf{Dataset} & \textbf{Classes} & \textbf{Description}\\
\hline
MNIST \cite{lecun_1998_cnn} & 10 (one class for each digit) & Counts 70000 grayscale images of handwritten digits of size $28\times28$ pixels.  \\
\hline
MNIST-M \cite{ganin_2016_da_dataset2} & 10 (one class for each digit) & \makecell{MNIST dataset extension for robustness evaluation with various transformations applied\\ to MNIST images, e.g., color jittering, rotation, and scaling. Contains 70000 color images of\\size $28\times28$ pixels.} \\
\hline
ImageNet-1K \cite{russakovsky_2015_imagenet} & 1000 (e.g., bird, cat, car). & \makecell{Contains 1.2 million color images of varying resolution showing objects from 1000 common\\ object categories.} \\
\hline 
SVHN \cite{netzer_2011_da_dataset3} & 10 (one class for each digit) & Contains 60000 color images of street view house numbers. Image size and quality vary. \\
\hline
USPS \cite{hull_1994_usps_dataset} & 10 (one class for each digit) & \makecell{Count 9298 grayscale images with handwritten digits of size $16 \times 16$ pixels. Image quality and\\orientations vary.}\\
\hline
CIFAR-10 \cite{coates_2011_da_dataset5} & 10 (e.g., airplane, bird, cat) & \makecell{Contains 60000 color images of size $32\times32$ of varying quality showing vehicles, animals, and\\other objects.}\\
\hline
STL-10 \cite{krizhevsky_2009_da_dataset4} & 10 (e.g., deer, fog, airplane) & \makecell{Contains 150000 color images of $96 \times 96$ of varying quality showing images with complex\\backgrounds, with 50000 images having labels.} \\
\hline
\end{tabular}
\end{center}
\end{table*}

\section{Discussion}
\label{sec:discussion}
\noindent\textbf{Domain generalization.} We distilled two common approaches for domain generalization, namely data augmentation, and DNN architectural modifications, including DNN depth and building blocks. We observe that there is no universal method to be applied in all scenarios, but that different methods are suitable for different situations. The literature on data augmentation agrees that the right choice of data augmentation can boost natural OOD robustness, but that multiple augmentations can boost robustness even more \cite{devries_2017_dg_aug_cutout, yang_2022_augmentations, hendrycks_2021_deepaugment}. However, the challenge remains in identifying the optimal augmentation set, which largely depends on the downstream task and dataset properties \cite{yang_2022_augmentations}. On the other hand, increasing network depth \cite{hendrycks_2018_benchmarking} and incorporating architectural building blocks that promote feature aggregation have been shown to enhance model robustness \cite{huang_2018_densenets, xie_2017_resnexe}. Increasing network depth depends on large and diverse training datasets to regularize the model, so this strategy is often combined with data augmentation.

\noindent \textbf{Domain adaptation} is often achieved through transfer learning, though other methods such as FSL and meta-learning have also been utilized for this purpose.  While transfer learning can enhance robustness on the target distribution, the method of knowledge transfer should be carefully selected based on the properties of source and target distributions. Both experimental and theoretical studies have shown that an inappropriate method can distort pretrained features learned from source distribution, exacerbating both ID and OOD generalization. Among supervised and SSL approaches to domain adaptation, SSL has demonstrated its superiority \cite{kumar_2022_finetuning}. As SSL continues to evolve toward new approaches for efficient representation learning, it can be expected that more domain adaptation methods will incorporate SSL. In particular, contrastive learning has demonstrated superior performance in representation learning field \cite{chen_2020_simclr, caron_2021_dino}. Learning better feature extractors can improve the linear probing approach to transfer learning or any hybrid method that includes linear probing. While a limited review of few-shot and meta-learning approaches suggests that they can improve robustness in domain adaptation, they may still be outperformed by simple transductive fine-tuning \cite{guneet_2020_fslbaseline}.

\section{Conclusions}
\label{sec:conclusions}

Since DL models are increasingly used in practice, a growing body of research focuses on minimizing the effect of data distribution shifts. In this review we have covered recent progress in domain generalization and domain adaptation techniques where the first approach aims to prepare the model for unknown data distribution shifts in training time and the second to adapt the model for known distribution shift in inference time. We also provide information on commonly used benchmarking datasets to provide the reader with better understanding of the reviewed methods.

\section*{Acknowledgment}
This paper has received funding from the European Union’s Horizon 2020 research and innovation programme under Grant Agreement number 85696. The paper has also been supported by the Ministry of Science, Technological Development, and Innovation of Republic of Serbia through project no. 451-03-47/2023-01/200156 “Innovative scientific and artistic research from the FTS domain”.

\bibliography{cite}

\begin{thebibliography}{10}
\providecommand{\url}[1]{#1}
\csname url@samestyle\endcsname
\providecommand{\newblock}{\relax}
\providecommand{\bibinfo}[2]{#2}
\providecommand{\BIBentrySTDinterwordspacing}{\spaceskip=0pt\relax}
\providecommand{\BIBentryALTinterwordstretchfactor}{4}
\providecommand{\BIBentryALTinterwordspacing}{\spaceskip=\fontdimen2\font plus
\BIBentryALTinterwordstretchfactor\fontdimen3\font minus
  \fontdimen4\font\relax}
\providecommand{\BIBforeignlanguage}[2]{{%
\expandafter\ifx\csname l@#1\endcsname\relax
\typeout{** WARNING: IEEEtran.bst: No hyphenation pattern has been}%
\typeout{** loaded for the language `#1'. Using the pattern for}%
\typeout{** the default language instead.}%
\else
\language=\csname l@#1\endcsname
\fi
#2}}
\providecommand{\BIBdecl}{\relax}
\BIBdecl

\bibitem{drenkow_2021_robustness}
\BIBentryALTinterwordspacing
N.~Drenkow, N.~Sani, I.~Shpitser, and M.~Unberath, ``Robustness in {Deep}
  {Learning} for {Computer} {Vision}: {Mind} the gap?'' Dec. 2021,
  arXiv:2112.00639 [cs]. [Online]. Available:
  \url{http://arxiv.org/abs/2112.00639}
\BIBentrySTDinterwordspacing

\bibitem{goodfellow_2014_adversarial}
I.~J. Goodfellow, J.~Shlens, and C.~Szegedy, ``Explaining and harnessing
  adversarial examples,'' \emph{arXiv preprint arXiv:1412.6572}, 2014.

\bibitem{sugiyama_2012_machine}
M.~Sugiyama and M.~Kawanabe, \emph{Machine learning in non-stationary
  environments: Introduction to covariate shift adaptation}.\hskip 1em plus
  0.5em minus 0.4em\relax MIT press, 2012.

\bibitem{hendrycks_2018_benchmarking}
D.~Hendrycks and T.~Dietterich, ``Benchmarking neural network robustness to
  common corruptions and perturbations,'' in \emph{Proc. ICLR}, 2019.

\bibitem{recht_2019_imagenet}
B.~Recht, R.~Roelofs, L.~Schmidt, and V.~Shankar, ``Do imagenet classifiers
  generalize to imagenet?'' in \emph{ICML}.\hskip 1em plus 0.5em minus
  0.4em\relax PMLR, 2019, pp. 5389--5400.

\bibitem{cifar10-dataset}
A.~Krizhevsky, ``{Learning Multiple Layers of Features from Tiny Images},''
  Master's thesis, University of Toronto, Canada, 2009.

\bibitem{imagenet-dataset}
J.~Deng, W.~Dong, R.~Socher, L.-J. Li, K.~Li, and L.~Fei-Fei, ``Imagenet: A
  large-scale hierarchical image database,'' in \emph{CVPR}, 2009, pp.
  248--255.

\bibitem{goodfellow_2016_dl}
I.~Goodfellow, Y.~Bengio, and A.~Courville, \emph{Deep Learning}.\hskip 1em
  plus 0.5em minus 0.4em\relax MIT Press, 2016,
  \url{http://www.deeplearningbook.org}.

\bibitem{kumar_2022_finetuning}
A.~Kumar, A.~Raghunathan, R.~M. Jones, T.~Ma, and P.~Liang, ``Fine-tuning can
  distort pretrained features and underperform out-of-distribution,'' in
  \emph{ICLR}, 2022.

\bibitem{croce_2020_advers1}
F.~Croce and M.~Hein, ``Reliable evaluation of adversarial robustness with an
  ensemble of diverse parameter-free attacks,'' in \emph{ICML}.\hskip 1em plus
  0.5em minus 0.4em\relax PMLR, 2020, pp. 2206--2216.

\bibitem{cohen_2019_adver2}
J.~Cohen, E.~Rosenfeld, and Z.~Kolter, ``Certified adversarial robustness via
  randomized smoothing,'' in \emph{ICML}.\hskip 1em plus 0.5em minus
  0.4em\relax PMLR, 2019, pp. 1310--1320.

\bibitem{madry_2018_adver3}
\BIBentryALTinterwordspacing
A.~Madry, A.~Makelov, L.~Schmidt, D.~Tsipras, and A.~Vladu, ``Towards deep
  learning models resistant to adversarial attacks,'' in \emph{ICLR}, 2018.
  [Online]. Available: \url{https://openreview.net/forum?id=rJzIBfZAb}
\BIBentrySTDinterwordspacing

\bibitem{hendrycks_2019_dg_augmix}
D.~Hendrycks, N.~Mu, E.~D. Cubuk, B.~Zoph, J.~Gilmer, and B.~Lakshminarayanan,
  ``Augmix: A simple data processing method to improve robustness and
  uncertainty,'' \emph{arXiv preprint arXiv:1912.02781}, 2019.

\bibitem{yun_2019_dg_aug_cutmix}
S.~Yun, D.~Han, S.~Chun, S.~J. Oh, Y.~Yoo, and J.~Choe, ``Cutmix:
  Regularization strategy to train strong classifiers with localizable
  features,'' in \emph{Proc. ICCV}, 2019, pp. 6022--6031.

\bibitem{devries_2017_dg_aug_cutout}
\BIBentryALTinterwordspacing
T.~Devries and G.~W. Taylor, ``Improved regularization of convolutional neural
  networks with cutout,'' \emph{CoRR}, vol. abs/1708.04552, 2017. [Online].
  Available: \url{http://arxiv.org/abs/1708.04552}
\BIBentrySTDinterwordspacing

\bibitem{hendrycks_2021_deepaugment}
\BIBentryALTinterwordspacing
D.~Hendrycks, S.~Basart, N.~Mu, S.~Kadavath, F.~Wang, E.~Dorundo, R.~Desai,
  T.~Zhu, S.~Parajuli, M.~Guo, D.~Song, J.~Steinhardt, and J.~Gilmer, ``The
  many faces of robustness: A critical analysis of out-of-distribution
  generalization,'' in \emph{ICCV}.\hskip 1em plus 0.5em minus 0.4em\relax Los
  Alamitos, CA, USA: IEEE Computer Society, oct 2021, pp. 8320--8329. [Online].
  Available:
  \url{https://doi.ieeecomputersociety.org/10.1109/ICCV48922.2021.00823}
\BIBentrySTDinterwordspacing

\bibitem{he_2016_deep_resnets}
K.~He, X.~Zhang, S.~Ren, and J.~Sun, ``Deep residual learning for image
  recognition,'' in \emph{Proc. CVPR}, 2016, pp. 770--778.

\bibitem{huang_2017_densenet}
G.~Huang, Z.~Liu, L.~Van Der~Maaten, and K.~Q. Weinberger, ``Densely connected
  convolutional networks,'' in \emph{Proc. CVPR}, 2017, pp. 4700--4708.

\bibitem{tan_2019_efficientnet}
M.~Tan and Q.~Le, ``Efficientnet: Rethinking model scaling for convolutional
  neural networks,'' in \emph{ICML}.\hskip 1em plus 0.5em minus 0.4em\relax
  PMLR, 2019, pp. 6105--6114.

\bibitem{dosovitskiy_2020_vit}
A.~Dosovitskiy, L.~Beyer, A.~Kolesnikov, D.~Weissenborn, X.~Zhai,
  T.~Unterthiner, M.~Dehghani, M.~Minderer, G.~Heigold, S.~Gelly \emph{et~al.},
  ``An image is worth 16x16 words: Transformers for image recognition at
  scale,'' \emph{arXiv preprint arXiv:2010.11929}, 2020.

\bibitem{rumelhart_1986_backprop}
\BIBentryALTinterwordspacing
D.~E. Rumelhart, G.~E. Hinton, and R.~J. Williams,
  ``\BIBforeignlanguage{en}{Learning representations by back-propagating
  errors},'' \emph{\BIBforeignlanguage{en}{Nature}}, vol. 323, no. 6088, pp.
  533--536, Oct. 1986, number: 6088 Publisher: Nature Publishing Group.
  [Online]. Available: \url{https://www.nature.com/articles/323533a0}
\BIBentrySTDinterwordspacing

\bibitem{duchi_2011_adagrad}
\BIBentryALTinterwordspacing
J.~Duchi, E.~Hazan, and Y.~Singer, ``Adaptive {Subgradient} {Methods} for
  {Online} {Learning} and {Stochastic} {Optimization},'' \emph{JMLR}, vol.~12,
  no.~61, pp. 2121--2159, 2011. [Online]. Available:
  \url{http://jmlr.org/papers/v12/duchi11a.html}
\BIBentrySTDinterwordspacing

\bibitem{tijmen_2012_rmsprop}
T.~Tieleman and G.~Hinton, ``Lecture 6.5 - rmsprop: Divide the gradient by a
  running average of its recent magnitude,'' \emph{Coursera: Neural Networks
  for Machine Learning}, 2012,
  \url{https://www.cs.toronto.edu/~tijmen/csc321/slides/lecture_slides_lec6.pdf}.

\bibitem{kingma_2015_adam}
D.~P. Kingma and J.~Ba, ``Adam: {A} method for stochastic optimization,'' in
  \emph{Proc. ICLR}, Y.~Bengio and Y.~LeCun, Eds., 2015.

\bibitem{bergstra_2013_hyperparam}
J.~Bergstra, D.~Yamins, and D.~D. Cox, ``Making a science of model search:
  Hyperparameter optimization in hundreds of dimensions for vision
  architectures,'' in \emph{Proc. ICML}, 2013.

\bibitem{hospedales_2022_metalearning}
T.~Hospedales, A.~Antoniou, P.~Micaelli, and A.~Storkey, ``Meta-learning in
  neural networks: A survey,'' \emph{PAMI}, vol.~44, no.~09, pp. 5149--5169,
  sep 2022.

\bibitem{srivastava_2014_dropout}
N.~Srivastava, G.~Hinton, A.~Krizhevsky, I.~Sutskever, and R.~Salakhutdinov,
  ``Dropout: a simple way to prevent neural networks from overfitting,''
  \emph{JMLR}, vol.~15, no.~1, pp. 1929--1958, 2014.

\bibitem{lecun_1998_cnn}
Y.~Lecun, L.~Bottou, Y.~Bengio, and P.~Haffner, ``Gradient-based learning
  applied to document recognition,'' \emph{Proceedings of the IEEE}, vol.~86,
  no.~11, pp. 2278--2324, 1998.

\bibitem{girshick_2014_rcnn}
R.~Girshick, J.~Donahue, T.~Darrell, and J.~Malik, ``Rich feature hierarchies
  for accurate object detection and semantic segmentation,'' in \emph{Proc.
  CVPR}, 2014, pp. 580--587.

\bibitem{ronneberger_2015_unet}
O.~Ronneberger, P.~Fischer, and T.~Brox, ``U-net: Convolutional networks for
  biomedical image segmentation,'' in \emph{MICCAI}.\hskip 1em plus 0.5em minus
  0.4em\relax Springer, 2015, pp. 234--241.

\bibitem{cheng_2021_instancesegm}
B.~Cheng, A.~Schwing, and A.~Kirillov, ``Per-pixel classification is not all
  you need for semantic segmentation,'' \emph{Advances in Neural Information
  Processing Systems}, vol.~34, pp. 17\,864--17\,875, 2021.

\bibitem{radford_2015_dcgan}
A.~Radford, L.~Metz, and S.~Chintala, ``Unsupervised representation learning
  with deep convolutional generative adversarial networks,'' \emph{arXiv
  preprint arXiv:1511.06434}, 2015.

\bibitem{dosovitskiy_2021_vit}
\BIBentryALTinterwordspacing
A.~Dosovitskiy, L.~Beyer, A.~Kolesnikov, D.~Weissenborn, X.~Zhai,
  T.~Unterthiner, M.~Dehghani, M.~Minderer, G.~Heigold, S.~Gelly, J.~Uszkoreit,
  and N.~Houlsby, ``An image is worth 16x16 words: Transformers for image
  recognition at scale,'' in \emph{ICLR}, 2021. [Online]. Available:
  \url{https://openreview.net/forum?id=YicbFdNTTy}
\BIBentrySTDinterwordspacing

\bibitem{yang_2022_augmentations}
S.~Yang, W.~Xiao, M.~Zhang, S.~Guo, J.~Zhao, and F.~Shen, ``Image data
  augmentation for deep learning: A survey,'' \emph{arXiv preprint
  arXiv:2204.08610}, 2022.

\bibitem{zhong_2017_dg_aug_erase}
\BIBentryALTinterwordspacing
Z.~Zhong, L.~Zheng, G.~Kang, S.~Li, and Y.~Yang, ``Random erasing data
  augmentation,'' \emph{CoRR}, vol. abs/1708.04896, 2017. [Online]. Available:
  \url{http://arxiv.org/abs/1708.04896}
\BIBentrySTDinterwordspacing

\bibitem{yun_2019_cutmix}
S.~Yun, D.~Han, S.~J. Oh, S.~Chun, J.~Choe, and Y.~Yoo, ``Cutmix:
  Regularization strategy to train strong classifiers with localizable
  features,'' in \emph{Proc. ICCV}, October 2019.

\bibitem{zhang_2018_mixup}
H.~Zhang, M.~Cisse, Y.~N. Dauphin, and D.~Lopez-Paz, ``mixup: Beyond empirical
  risk minimization,'' in \emph{ICLR}, 2018.

\bibitem{taori_2020_measuring}
R.~Taori, A.~Dave, V.~Shankar, N.~Carlini, B.~Recht, and L.~Schmidt,
  ``Measuring {Robustness} to {Natural} {Distribution} {Shifts} in {Image}
  {Classification},'' in \emph{NeurIPS}, H.~Larochelle, M.~Ranzato, R.~Hadsell,
  M.~F. Balcan, and H.~Lin, Eds., vol.~33.\hskip 1em plus 0.5em minus
  0.4em\relax Curran Associates, Inc., 2020, pp. 18\,583--18\,599.

\bibitem{goodfellow_2014_gans}
I.~Goodfellow, J.~Pouget-Abadie, M.~Mirza, B.~Xu, D.~Warde-Farley, S.~Ozair,
  A.~Courville, and Y.~Bengio, ``Generative {Adversarial} {Nets},'' in
  \emph{NIPS}, Z.~Ghahramani, M.~Welling, C.~Cortes, N.~Lawrence, and K.~Q.
  Weinberger, Eds., vol.~27.\hskip 1em plus 0.5em minus 0.4em\relax Curran
  Associates, Inc., 2014.

\bibitem{cubuk_2019_autoaugment}
E.~D. Cubuk, B.~Zoph, D.~Mane, V.~Vasudevan, and Q.~V. Le, ``Autoaugment:
  Learning augmentation strategies from data,'' in \emph{Proc. CVPR}, 2019, pp.
  113--123.

\bibitem{lim_2019_fastautoaugment}
S.~Lim, I.~Kim, T.~Kim, C.~Kim, and S.~Kim, ``Fast autoaugment,'' \emph{Adv
  Neural Inf Process Syst}, vol.~32, 2019.

\bibitem{ho_2019_population}
D.~Ho, E.~Liang, X.~Chen, I.~Stoica, and P.~Abbeel, ``Population based
  augmentation: Efficient learning of augmentation policy schedules,'' in
  \emph{ICML}.\hskip 1em plus 0.5em minus 0.4em\relax PMLR, 2019, pp.
  2731--2741.

\bibitem{cubuk_2020_randaugment}
E.~D. Cubuk, B.~Zoph, J.~Shlens, and Q.~V. Le, ``Randaugment: Practical
  automated data augmentation with a reduced search space,'' in \emph{Proc.
  CVPRW}, 2020, pp. 702--703.

\bibitem{huang_2018_densenets}
\BIBentryALTinterwordspacing
G.~Huang, Z.~Liu, L.~van~der Maaten, and K.~Q. Weinberger, ``Densely
  {Connected} {Convolutional} {Networks},'' Jan. 2018. [Online]. Available:
  \url{http://arxiv.org/abs/1608.06993}
\BIBentrySTDinterwordspacing

\bibitem{xie_2017_resnexe}
S.~Xie, R.~Girshick, P.~Doll{\'a}r, Z.~Tu, and K.~He, ``Aggregated residual
  transformations for deep neural networks,'' in \emph{Proc. ICPR}, 2017, pp.
  1492--1500.

\bibitem{xie_2019_intriguing}
C.~Xie and A.~Yuille, ``Intriguing properties of adversarial training at
  scale,'' \emph{arXiv preprint arXiv:1906.03787}, 2019.

\bibitem{ke_2017_multigrid}
T.-W. Ke, M.~Maire, and S.~X. Yu, ``Multigrid neural architectures,'' in
  \emph{Proc. CVPR}, 2017, pp. 6665--6673.

\bibitem{huang_2018_multiscale}
G.~Huang, D.~Chen, T.~Li, F.~Wu, L.~van~der Maaten, and K.~Weinberger,
  ``Multi-scale dense networks for resource efficient image classification,''
  in \emph{ICLR}, 2018.

\bibitem{csurka_2017_domain_adaptation}
G.~Csurka \emph{et~al.}, \emph{Domain adaptation in computer vision
  applications}.\hskip 1em plus 0.5em minus 0.4em\relax Springer, 2017.

\bibitem{pan_2010_transferlearning}
S.~J. Pan and Q.~Yang, ``A survey on transfer learning,'' \emph{IEEE Trans.
  Knowl. Data Eng.}, vol.~22, no.~10, pp. 1345--1359, 2010.

\bibitem{wang_2020_fewshotlearning}
Y.~Wang, Q.~Yao, J.~T. Kwok, and L.~M. Ni, ``Generalizing from a few examples:
  A survey on few-shot learning,'' \emph{ACM computing surveys (csur)},
  vol.~53, no.~3, pp. 1--34, 2020.

\bibitem{he_2019_rethinking}
K.~He, R.~Girshick, and P.~Doll{\'a}r, ``Rethinking imagenet pre-training,'' in
  \emph{Proc. ICCV}, 2019, pp. 4918--4927.

\bibitem{yosinski_2014_transferable}
J.~Yosinski, J.~Clune, Y.~Bengio, and H.~Lipson, ``How transferable are
  features in deep neural networks?'' \emph{NIPS}, vol.~27, 2014.

\bibitem{hendrycks_2019_robustness_pretraining}
D.~Hendrycks, K.~Lee, and M.~Mazeika, ``Using pre-training can improve model
  robustness and uncertainty,'' in \emph{ICML}.\hskip 1em plus 0.5em minus
  0.4em\relax PMLR, 2019, pp. 2712--2721.

\bibitem{sun_2019_unsupervised_da}
Y.~Sun, E.~Tzeng, T.~Darrell, and A.~A. Efros, ``Unsupervised domain adaptation
  through self-supervision,'' \emph{arXiv preprint arXiv:1909.11825}, 2019.

\bibitem{yu_2020_test-time}
Y.~Sun, X.~Wang, Z.~Liu, J.~Miller, A.~Efros, and M.~Hardt, ``Test-time
  training with self-supervision for generalization under distribution
  shifts,'' in \emph{Proc. ICML}, ser. PMLR, H.~D. III and A.~Singh, Eds., vol.
  119.\hskip 1em plus 0.5em minus 0.4em\relax PMLR, 13--18 Jul 2020, pp.
  9229--9248.

\bibitem{feng_2019_sslrotation}
Z.~Feng, C.~Xu, and D.~Tao, ``Self-supervised representation learning by
  rotation feature decoupling,'' in \emph{Proc. CVPR}, 2019, pp.
  10\,364--10\,374.

\bibitem{chen_2020_simclr}
T.~Chen, S.~Kornblith, M.~Norouzi, and G.~Hinton, ``A simple framework for
  contrastive learning of visual representations,'' in \emph{ICML}.\hskip 1em
  plus 0.5em minus 0.4em\relax PMLR, 2020, pp. 1597--1607.

\bibitem{he_2020_moco}
K.~He, H.~Fan, Y.~Wu, S.~Xie, and R.~Girshick, ``Momentum contrast for
  unsupervised visual representation learning,'' in \emph{Proc. CVPR}, 2020,
  pp. 9729--9738.

\bibitem{caron_2021_dino}
M.~Caron, H.~Touvron, I.~Misra, H.~J{\'e}gou, J.~Mairal, P.~Bojanowski, and
  A.~Joulin, ``Emerging properties in self-supervised vision transformers,'' in
  \emph{Proc. CVPR}, 2021, pp. 9650--9660.

\bibitem{caruana_1998_multitask}
R.~Caruana, \emph{Multitask learning}.\hskip 1em plus 0.5em minus 0.4em\relax
  Springer, 1998.

\bibitem{ravi_2017_optimization}
S.~Ravi and H.~Larochelle, ``Optimization as a model for few-shot learning,''
  in \emph{ICLR}, 2017.

\bibitem{finn_2017_metalearning_da}
C.~Finn, P.~Abbeel, and S.~Levine, ``Model-agnostic meta-learning for fast
  adaptation of deep networks,'' in \emph{ICML}.\hskip 1em plus 0.5em minus
  0.4em\relax PMLR, 2017, pp. 1126--1135.

\bibitem{li_2020_metalearning_da}
D.~Li and T.~Hospedales, ``Online meta-learning for multi-source and
  semi-supervised domain adaptation,'' in \emph{ECCV}, A.~Vedaldi, H.~Bischof,
  T.~Brox, and J.-M. Frahm, Eds.\hskip 1em plus 0.5em minus 0.4em\relax Cham:
  Springer International Publishing, 2020, pp. 382--403.

\bibitem{saito_2020_universal}
K.~Saito, D.~Kim, S.~Sclaroff, and K.~Saenko, ``Universal domain adaptation
  through self-supervision,'' \emph{NIPS}, vol.~33, pp. 16\,282--16\,292, 2020.

\bibitem{tseng_2020_crossdomain_fsl_da}
H.-Y. Tseng, H.-Y. Lee, J.-B. Huang, and M.-H. Yang, ``Cross-domain few-shot
  classification via learned feature-wise transformation,'' \emph{arXiv
  preprint arXiv:2001.08735}, 2020.

\bibitem{guo_2020_broader}
Y.~Guo, N.~C. Codella, L.~Karlinsky, J.~V. Codella, J.~R. Smith, K.~Saenko,
  T.~Rosing, and R.~Feris, ``A broader study of cross-domain few-shot
  learning,'' in \emph{ECCV}.\hskip 1em plus 0.5em minus 0.4em\relax Springer,
  2020, pp. 124--141.

\bibitem{ge_2019_deepfashion2}
Y.~Ge, R.~Zhang, X.~Wang, X.~Tang, and P.~Luo, ``Deepfashion2: A versatile
  benchmark for detection, pose estimation, segmentation and re-identification
  of clothing images,'' in \emph{Proc. CVPR}, 2019, pp. 5337--5345.

\bibitem{shu_2018_dirt}
R.~Shu, H.~H. Bui, H.~Narui, and S.~Ermon, ``A dirt-t approach to unsupervised
  domain adaptation,'' \emph{arXiv preprint arXiv:1802.08735}, 2018.

\bibitem{ganin_2016_da_dataset2}
Y.~Ganin, E.~Ustinova, H.~Ajakan, P.~Germain, H.~Larochelle, F.~Laviolette,
  M.~Marchand, and V.~Lempitsky, ``Domain-adversarial training of neural
  networks,'' \emph{J Mach Learn Res}, vol.~17, no.~1, pp. 2096--2030, 2016.

\bibitem{russakovsky_2015_imagenet}
O.~Russakovsky, J.~Deng, H.~Su, J.~Krause, S.~Satheesh, S.~Ma, Z.~Huang,
  A.~Karpathy, A.~Khosla, M.~Bernstein \emph{et~al.}, ``Imagenet large scale
  visual recognition challenge,'' \emph{Int J Comput Vis}, vol. 115, pp.
  211--252, 2015.

\bibitem{netzer_2011_da_dataset3}
Y.~Netzer, T.~Wang, A.~Coates, A.~Bissacco, B.~Wu, and A.~Y. Ng, ``Reading
  digits in natural images with unsupervised feature learning,'' 2011.

\bibitem{hull_1994_usps_dataset}
J.~Hull, ``A database for handwritten text recognition research,'' \emph{PAMI},
  vol.~16, no.~5, pp. 550--554, 1994.

\bibitem{krizhevsky_2009_da_dataset4}
A.~Krizhevsky, G.~Hinton \emph{et~al.}, ``Learning multiple layers of features
  from tiny images,'' 2009.

\bibitem{coates_2011_da_dataset5}
\BIBentryALTinterwordspacing
A.~Coates, A.~Ng, and H.~Lee, ``An analysis of single-layer networks in
  unsupervised feature learning,'' in \emph{Proc. AISTATS}, ser. PMLR,
  G.~Gordon, D.~Dunson, and M.~Dudík, Eds., vol.~15.\hskip 1em plus 0.5em
  minus 0.4em\relax Fort Lauderdale, FL, USA: PMLR, 11--13 Apr 2011, pp.
  215--223. [Online]. Available:
  \url{https://proceedings.mlr.press/v15/coates11a.html}
\BIBentrySTDinterwordspacing

\bibitem{hendrycks_2021_imageneta}
D.~Hendrycks, K.~Zhao, S.~Basart, J.~Steinhardt, and D.~Song, ``Natural
  adversarial examples,'' in \emph{Proc. CVPR}, 2021, pp. 15\,262--15\,271.

\bibitem{peng_2019_domainnet}
X.~Peng, Q.~Bai, X.~Xia, Z.~Huang, K.~Saenko, and B.~Wang, ``Moment matching
  for multi-source domain adaptation,'' in \emph{Proc. ICCV}, 2019, pp.
  1406--1415.

\bibitem{guneet_2020_fslbaseline}
\BIBentryALTinterwordspacing
G.~S. Dhillon, P.~Chaudhari, A.~Ravichandran, and S.~Soatto, ``A baseline for
  few-shot image classification,'' in \emph{ICLR}, 2020. [Online]. Available:
  \url{https://openreview.net/forum?id=rylXBkrYDS}
\BIBentrySTDinterwordspacing

\end{thebibliography}
\bibliographystyle{IEEEtran}

\vfill

\end{document}